\documentclass[10pt,twocolumn,letterpaper]{article}

\usepackage{cvpr}              
\usepackage{multirow}
\usepackage{balance}
\providecommand{\keywords}[1]
{
  \textbf{\textit{Keywords---}} #1
}
\definecolor{cvprblue}{rgb}{0.21,0.49,0.74}
\usepackage[pagebackref,breaklinks,colorlinks,allcolors=cvprblue]{hyperref}

\def\paperID{21} 
\def\confName{CVPR}
\def\confYear{2025}

\title{Task-conditioned Ensemble of Expert Models for Continuous Learning}

\author{Renu Sharma,  Debasmita Pal,  Arun Ross \\
Michigan State University, USA\\
{\tt\small sharma90@msu.edu, paldebas@msu.edu, rossarun@cse.msu.edu}
}

\usepackage{fancyhdr}
\fancypagestyle{firstpage}
{   
    \fancyhead[L]{}
    \fancyhead[R]{}
    \fancyhead[C]{\textcolor{red}{\textbf{IEEE/CVF Conference on Computer Vision and Pattern Recognition Workshops (CVPRW), Nashville, USA, 2025}}}
}

\begin{document}
\maketitle
\begin{abstract}
One of the major challenges in machine learning is maintaining the accuracy of the deployed model (e.g., a classifier) in a non-stationary environment. The non-stationary environment results in distribution shifts and, consequently, a degradation in accuracy. Continuous learning of the deployed model with new data could be one remedy. However, the question arises as to how we should update the model with new training data so that it retains its accuracy on the old data while adapting to the new data. In this work, we propose a task-conditioned ensemble of models to maintain the performance of the existing model. The method involves an ensemble of expert models based on task membership information. The in-domain models—based on the local outlier concept (different from the expert models) provide task membership information dynamically at run-time to each probe sample. To evaluate the proposed method, we experiment with three setups: the first represents distribution shift between tasks (LivDet-Iris-2017), the second represents distribution shift both between and within tasks (LivDet-Iris-2020), and the third represents disjoint distribution between tasks (Split MNIST). The experiments highlight the benefits of the proposed method. The source code is available at \href{https://github.com/iPRoBe-lab/Continuous_Learning_FE_DM/tree/main}{GitHub}.
\end{abstract}

\thispagestyle{firstpage}
\section{Introduction}
\label{sec:Introduction}

While much of the research in machine learning focuses on achieving higher accuracy in various classification and regression tasks, maintaining that level of performance in non-stationary environments \cite{Moreno-Torres2012} remains a relatively underexplored area. Non-stationary environments arise from factors such as changes in sensors, location, population groups, and other variables. These changes lead to distribution shifts—shifts in input and output distributions or their relationships—that degrade model performance \cite{Delange2021, Li2017, Moreno-Torres2012}. To sustain the performance of deployed models, it is essential to enable continuous learning with new task data while retaining knowledge from previous tasks.

One straightforward solution for continuous learning is to fine-tune the model with new data. However, this often leads to catastrophic forgetting of previously learned tasks \cite{Delange2021, Zhiqiang2023,Li2017}. Another option is to retrain the model using a comprehensive dataset that includes both old and new data. However, in real-world applications, old training data is often unavailable due to security, privacy, or operational constraints. Several continuous learning approaches \cite{Wang2024}, such as regularization-based \cite{Aljundi2018, Kao2021}, replay-based \cite{Bang2021, Ibrahim2024}, optimization-based \cite{Mehrdad2020, Guoliang2021}, representation-based \cite{Pham2021, Wu2022}, and architecture-based \cite{Xue2022, Douillard2022} methods, have been proposed in the literature. Most of these strategies aim to learn all subsequent tasks using a shared set of parameters (i.e., a single model), which can lead to significant inter-task interference \cite{Doan2023,Ramesh2022,Marouf2024}. This challenge becomes even more pronounced as the number of tasks increases, placing the entire burden on a single model.

We propose a multi-model approach in which each model is responsible for a specific task (an expert model), and the final score is obtained by dynamically merging the scores  of these expert models based on their task membership information. We consider a practical scenario where task information is not explicitly provided, but sufficient task-specific data is available to either train an expert model or utilize an already available expert model. Unlike previous methods \cite{Ibrahim2024, Guoliang2021, Kao2021, Xue2022}, which often involve updating the training data, strategies, or architecture of expert models, our approach does not interfere with the training process of the expert models. Instead, we offer a framework that enables the reuse of existing expert models, each tailored to a particular task. Our main contributions are as follows:

\noindent 1. We propose an ensemble of expert models, where the final prediction is obtained by combining scores from individual expert models based on task membership information.

\noindent 2. We introduce an in-domain model that dynamically estimates task membership information for each probe sample, using the concept of local outlier detection.

\noindent 3. We validate the effectiveness of our method through experiments on three diverse setups, each capturing different types of distribution shifts: LivDet-Iris-2017 (distribution shifts between tasks), LivDet-Iris-2020 (distribution shifts both between and within tasks), and Split MNIST (disjoint distributions between tasks).

\section{Related Work}
\label{sec:RelatedWork}


Continuous learning encompasses three scenarios \cite{Hsu2018, Vandeven2018}: Task-Incremental Learning (Task-IL), Domain-Incremental Learning (Domain-IL), and Class-Incremental Learning (Class-IL). Task-IL involves incrementally learning several independent tasks, with explicit knowledge of the task identity. Domain-IL focuses on learning tasks from the same output class-label space, but with varying input distributions, and without task identity. Class-IL, on the other hand, involves incrementally learning new classes in each task, without any information about the task identity. In this work, we focus on the Domain-IL scenario for continuous learning.

In the literature, continuous learning methodologies are typically categorized into five main groups \cite{Wang2024}: regularization-based, replay-based, optimization-based, representation-based, and architecture-based approaches. Regularization-based approaches introduce additional constraints to the learning process, penalizing drastic changes in model parameters \cite{Aljundi2018, Kao2021, Kirkpatrick2017, Li2017, Ritter2018, Schwarz2018, Zeng2019, Zenke2017}. Replay-based approaches enhance existing expert models with memory mechanisms to retain information about previously learned tasks \cite{Bang2021, Chaudhry2019, Ibrahim2024, Lopez2017, Prabhu2020, Shin2017, Van2018, Wu2019}. Optimization-based techniques focus on explicitly designing or manipulating optimization algorithms, such as using gradient projection in the gradient or input space of old tasks \cite{Mehrdad2020}, meta-learning for sequentially arriving tasks within the inner loop, and exploiting mode connectivity and flat minima in the loss landscape \cite{Guoliang2021, Mirzadeh2020b}. Representation-based approaches leverage the strengths of learned representations, including sparse representations from meta-training \cite{Javed2019}, self-supervised learning (SSL) \cite{Pham2021, Madaan2021}, and large-scale pre-training \cite{Wu2022, Shi2022}. Architecture-based approaches involve task-specific or adaptive parameters within a well-designed architecture, such as assigning dedicated parameters to each task (parameter allocation) \cite{Xue2022, Kang2022}, constructing task-adaptive sub-modules or subnetworks (modular networks) \cite{Rusu2016, Rajasegaran2019}, and decomposing the model into task-sharing and task-specific components (model decomposition) \cite{Wu2021, Douillard2022}.

In the model decomposition category of architecture-based approaches, ensembles or multiple networks are used for continuous learning. 
Doan et al. \cite{Doan2023} utilized ensembles of models and made them computationally efficient by leveraging neural network subspaces. 
The Model Zoo approach \cite{Ramesh2022}, inspired by boosting, grows an ensemble of small models, each trained during one episode of continuous learning. CoSCL \cite{Wang2022b} fixed the number of narrower sub-networks to learn all incremental tasks in parallel and encouraged cooperation among them by penalizing the differences in predictions made by their feature representations. MERLIN \cite{Joseph2020} assumed that the weights of a neural network for solving any task come from a meta-distribution, which they learned incrementally. 
The CoMA \cite{Marouf2024} method selectively weights each parameter in the weight ensemble by leveraging the Fisher information of the model's weights. Lee et al. \cite{Lee2020} proposed a closely related approach, where they learned separate expert models for each task and measured the marginal likelihood of the expert models using a density estimator. In contrast, our method dynamically assigns task membership information to expert models using an in-domain model.

\section{Proposed Approach}
\label{sec:Algorithm}

\begin{figure*}[h]
\centering
\includegraphics[scale=0.35]{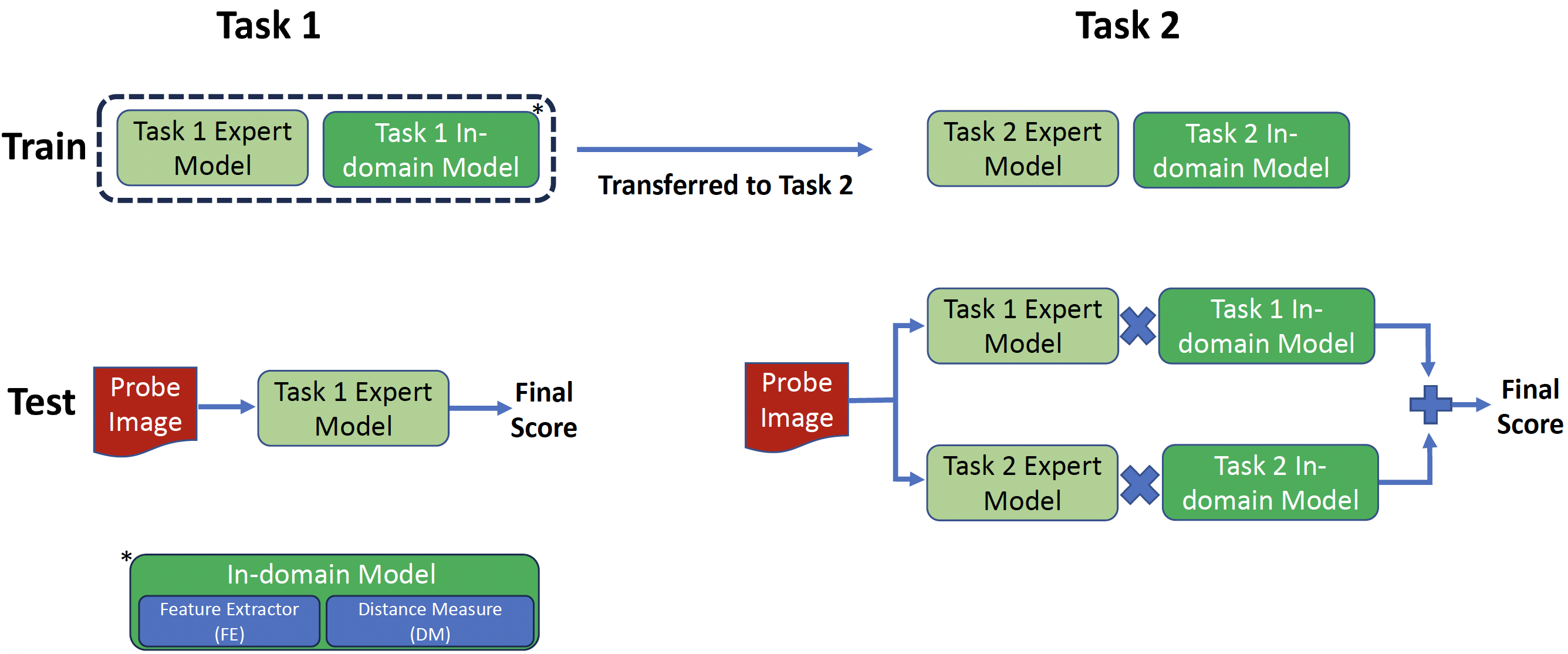}
\caption{The overall idea of the task-conditioned ensemble of models of continuous learning. Task 1 inference utilizes its only expert model, whereas task 2 inference utilizes both expert models with the help of in-domain models that provide membership information. In-domain model consists of two components: Feature Extractor (FE) and Distance Measure (DM).}
\label{fig:Overview}
\vspace{-10pt}
\end{figure*}

In this work, we propose a task-conditioned ensemble framework of expert models that preserves performance across all learned tasks. Figure \ref{fig:Overview} illustrates the proposed approach. Consider a scenario with two tasks (Task 1 and Task 2). For Task 1, we use its corresponding expert model for inference, while simultaneously training an in-domain model. For Task 2, we train separate expert and in-domain models specific to Task 2's training data. During inference, we combine the predictions from the expert models using their respective in-domain models. These in-domain models provide membership information for the probe sample, indicating the task it belongs to. Importantly, probe samples can come from any task, without revealing the task identity. In other words, the in-domain model estimates the task identity of a probe sample. The final prediction score for Task 2 is calculated as follows:
\begin{equation}
s = s_1.m_1 + s_2.m_2
\end{equation}
where, $s_1$ and $s_2$ are prediction scores from expert models, and $m_1$ and $m_2$ are membership scores from in-domain models.

Our main contribution lies in the introduction of the in-domain model, which estimates membership scores. The in-domain model operates on the principle of outlier detection: for each probe sample, it determines the degree to which the sample is an outlier with respect to the training distributions. To achieve this, the in-domain model consists of two components: (i) a \textit{feature extractor} that represents task-specific data in feature space, and (ii) a \textit{distance measure} that provides a membership score based on the outlier score of the probe sample relative to the task-specific feature space. The details of these components are as follows.

\textbf{Feature Extractor (FE):} The base architecture we use for feature extraction is a Vision Transformer (ViT) \cite{Dosovitskiy2021}. The success of transformers in natural language processing \cite{Vaswani2017} and computer vision \cite{Dosovitskiy2021} inspired us to adopt them for representing task-specific data. While convolutional neural networks (CNNs) capture local structures in images, they are less effective at modeling global information. In contrast, ViT excels at modeling global contextual information through its self-attention mechanism, making it highly suited for representing task-level information \cite{Maurício2023}. 
We train the ViT feature extractor using two losses: the center loss and the mean-shifted intra-class loss. The details of these losses are as follows:

\textbf{1. Center Loss:} The objective of the center loss is to extract features from the training data such that all feature embeddings get closer to the center of the embeddings. The center of the training data embeddings is calculated as 
 \begin{equation}
     c= \mathbb{E}_{x\in \chi_{train}} [\phi(x)]
 \end{equation}
where, $x$ is the input image, $\phi(x) $ is the feature embedding from the ViT model, and $\chi_{train}$ is the train set. The $c$ is initialized with pre-trained ViT features. Thereafter, it gets updated in each epoch. The center loss is then calculated as 
 \begin{equation}
 \ell_{center} (x) = \lVert \phi(x) -c \rVert^2.
 \end{equation}
The loss reduces the intra-train set variations among training data and forms a closer feature space for the samples. This helps in detecting outlier samples in presence of other task data.

\begin{figure}[t]
\centering
\includegraphics[scale=0.25]{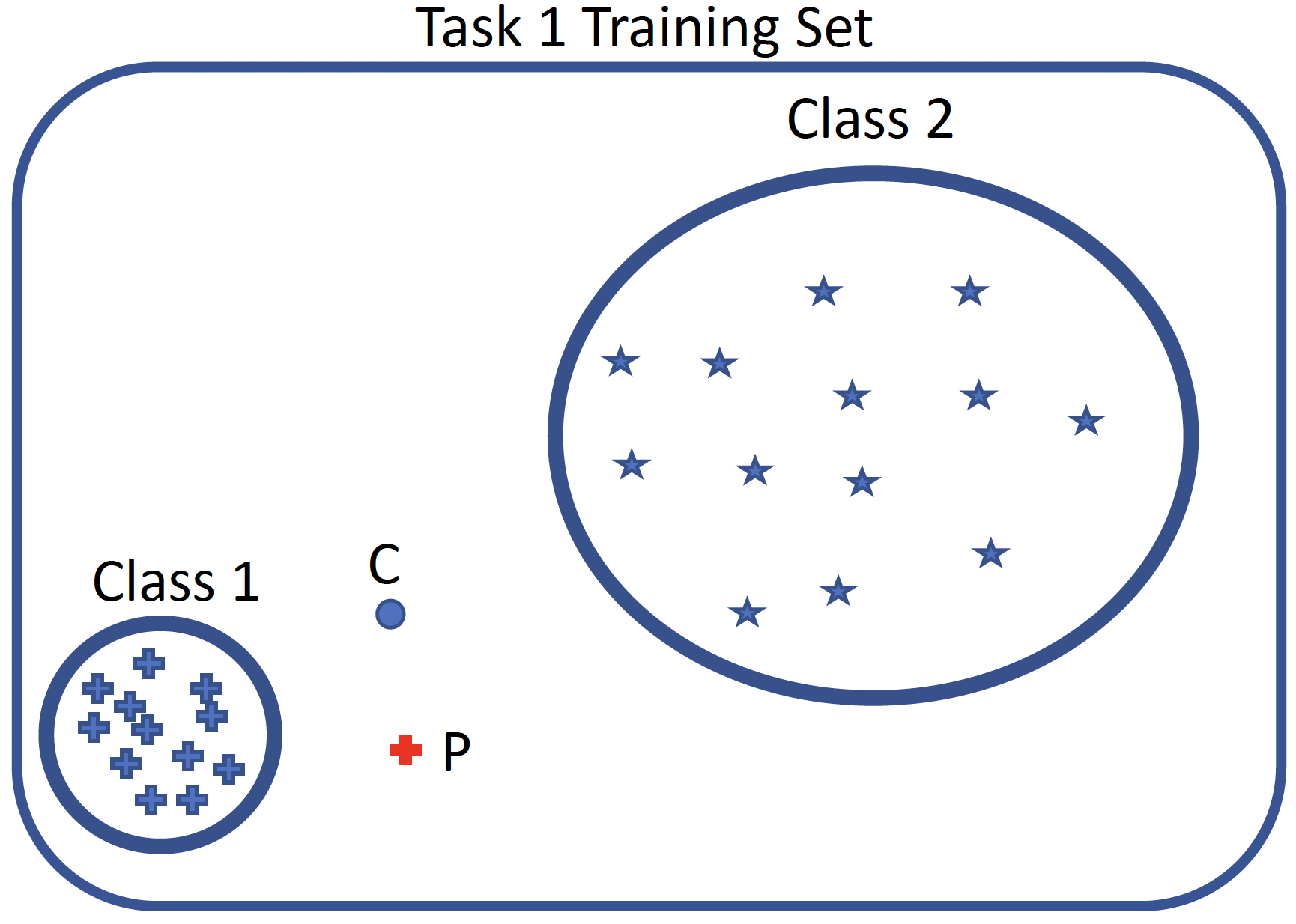}
\caption{Illustration of a local outlier concept, motivation for defining feature space. Blue-colored data points belong to one training set; C is the center of the training set; and red-colored data point P is a probe sample. There are two classes (Class 1 and Class 2) in the blue-colored training set. If we consider the global outlier concept, the red-colored probe sample would be an inlier. However, if the local outlier concept is used, the probe sample is an outlier to both Class 1 and Class 2 as well as to the blue-colored training set. The figure is better viewed in color.}
\label{fig:LocalOutlier}
\vspace{-10pt}
\end{figure}

\textbf{2. Mean-Shifted Intra-Class Loss:} The objective of this loss is to form a cluster of samples belonging to the same class around the center of feature embeddings. To accomplish the objective, we first mean-shifted the embeddings of the training samples as
\begin{equation}
    \theta(x) = \frac{ \phi(x) -c }{\lVert \phi(x) -c \rVert^2}
\end{equation}
where, $\phi(x)$ is the feature embedding of input sample $x$ and $c$ is the center of the feature embeddings. We then estimate contrastive loss over the  mean-shifted embeddings. Let $x_{i1}$ and $x_{i2}$ be two input images belonging to the same class $C_i$, then loss is defined as:
\begin{multline}
     \ell_{msic} (x_{i1},x_{i2})_{\{x_{i1},x_{i2}\} \in C_i} = \ell_{con} (\theta(x_{i1}),\theta(x_{i2})) \\
    = -\log \frac{\exp((\theta(x_{i1}).\theta(x_{i2}))/\tau)}{\sum_{j \neq i}^{2N} \exp((\theta(x_{i1}).\theta(x_j))/\tau)}
\end{multline}
where, $\theta(.)$ is the mean-shifted embedding, $N$ is the batch size, and $\tau$ is the temperature hyperparameter. Together, the two losses create a feature space where samples from the same class form a cluster close to the center of the training data. The class cluster formation helps in the detection of local outliers. By local outlier, we refer to a sample that has a low distance to the center of the training set but is an outlier with respect to other class distributions. Consider Figure \ref{fig:LocalOutlier}, where the blue-colored data points represent the training set, C is the center of the training set, and the red-colored point P is a probe sample. There are two classes (Class 1 and Class 2) with different densities in the training set. If we consider the global outlier concept (measured by the distance to the center of the training set), the probe sample P would be an inlier to the blue-colored training set since its distance to the center is smaller compared to the data points of Class 1 and Class 2. However, according to the concept of local outlier, P is an outlier with respect to both Class 1 and Class 2 because the distance between data points within Class 1 or Class 2 is smaller than the distance between the probe sample and the data points of Class 1 or Class 2. As a result, the probe sample P is considered an outlier with respect to the entire blue-colored training set. Therefore, the local outlier concept is more effective for membership assignment because it focuses not only on the overall task distribution but also on the class distribution within each task.

The total loss used to train the feature extractor is the sum of center loss and mean-shifted intra-class loss: 
\begin{equation}
    \ell_{total} (x',x'')  = \ell_{center}(x') + \ell_{center}(x'') + \ell_{msic}(x',x'').
\end{equation}

\textbf{Distance Measure (DM)}: After representing the training data and the probe sample, we estimate the distance of the probe sample with respect to the training data using the Local Outlier Factor (LOF) \cite{Breunig2000}. LOF is a density-based local outlier detection technique that identifies anomalous points relative to a local cluster of neighboring points by incorporating a nearest-neighbor algorithm. It detects outliers whose density is significantly lower than that of their neighbors. The losses we proposed to represent the training data help in estimating the local outliers using LOF. If the LOF score is approximately 1, it suggests that the sample has a density similar to that of its neighbors. A score less than 1 indicates that the sample has a higher local density than its neighbors, while a score greater than 1 indicates that the sample has a lower density than its neighbors. To assign a membership score to each expert model, we first invert the LOF scores ($l_1, l_2$) and then apply SoftMax normalization, as follows:
\begin{equation}
    (m_1, m_2) = softmax\left(\frac{1}{l_1},\frac{1}{l_2}\right).
\end{equation}

\section{Experimental Setup and Results}
\label{sec:ExpSetupResults}

\begin{table*}[h!]
\captionsetup{font=small}
\caption{Description of the task 1 and 2 training/test sets in the LivDet-Iris-2017 setup along with the number of bonafide and fake iris images present in the datasets. Each test set represents different testing scenarios. Here, “K. Test” means a known test set of the dataset, and “U. Test” means an unknown test set as defined in \cite{Yambay2017}}
\label{table:LivDet-Iris-2017-Dataset}
\resizebox{\textwidth}{!}{%
\begin{tabular}{|l|cc|cccccccccc|}
\hline
\textbf{\large Domains} & \multicolumn{2}{c|}{\textbf{\begin{tabular}[c]{@{}c@{}}\large Task 1 Train and Test Sets \\ \large (Proprietary Dataset)\end{tabular}}} & \multicolumn{10}{c|}{\textbf{\large Task 2 Train and Test Sets  (LivDet-Iris-2017 Dataset)}} \\ \hline
\textbf{\large Datasets} & \multicolumn{1}{c|}{\textbf{\begin{tabular}[c]{@{}c@{}} \large Proprietary  Split I \end{tabular}}} & \textbf{\begin{tabular}[c]{@{}c@{}}\large Proprietary Split II\end{tabular}} & \multicolumn{2}{c|}{\textbf{\begin{tabular}[c]{@{}c@{}} \large Clarkson\end{tabular}}} & \multicolumn{3}{c|}{\textbf{\begin{tabular}[c]{@{}c@{}} \large Warsaw\end{tabular}}} & \multicolumn{3}{c|}{\textbf{\begin{tabular}[c]{@{}c@{}} \large Notre Dame\end{tabular}}} & \multicolumn{2}{c|}{\textbf{\begin{tabular}[c]{@{}c@{}} \large IIITD-WVU\end{tabular}}} \\ \hline
\textbf{\large Train/Test} & \multicolumn{1}{c|}{\textbf{\large Train}} & \textbf{\large Test} & \multicolumn{1}{c|}{\textbf{\large Train}} & \multicolumn{1}{c|}{\textbf{\large Test}} & \multicolumn{1}{c|}{\textbf{\large Train}} & \multicolumn{1}{c|}{\textbf{\large K. Test}} & \multicolumn{1}{c|}{\textbf{\large U. Test}} & \multicolumn{1}{c|}{\textbf{\large Train}} & \multicolumn{1}{c|}{\textbf{\large K. Test}} & \multicolumn{1}{c|}{\textbf{\large U. Test}} & \multicolumn{1}{c|}{\textbf{\large Train}} & \textbf{\large Test} \\ \hline
\large Bonafide & \multicolumn{1}{c|}{\large 9,660} & \large 2,963 & \multicolumn{1}{c|}{\large 2,469} & \multicolumn{1}{c|}{ \large 1,485} & \multicolumn{1}{c|}{\large 1,844} & \multicolumn{1}{c|}{\large 974} & \multicolumn{1}{c|}{\large 2,350} & \multicolumn{1}{c|}{ \large 600} & \multicolumn{1}{c|}{\large 900} & \multicolumn{1}{c|}{\large 900} & \multicolumn{1}{c|}{ \large 2,250} & \large 702 \\ \hline
\large PA & \multicolumn{1}{c|}{\large 6,075} & \multicolumn{1}{c|}{\large 352} & \multicolumn{1}{c|}{\large 2,468} & \multicolumn{1}{c|}{\large 1,673} & \multicolumn{1}{c|}{\large 2,669} & \multicolumn{1}{c|}{\large 2,016} & \multicolumn{1}{c|}{\large 2,160} & \multicolumn{1}{c|}{\large 600} & \multicolumn{1}{c|}{\large 900} & \multicolumn{1}{c|}{\large 900} & \multicolumn{1}{c|}{\large 4,000} & \large 3,507 \\ \hline
\end{tabular}
}
\end{table*}

\begin{table*}[]
\captionsetup{font=small}
\caption{The performance of all methods in terms of True Detection Rate (\%, higher the better) at 0.2\% False Detection Rate on task 1 and 2 test sets of the LivDet-Iris-2017 setup.}
\label{table:LivDet-Iris-2017-Results}
\resizebox{\textwidth}{!}{%
\begin{tabular}{|lcclcccccccc|}
\hline
\multicolumn{1}{|l|}{\textbf{Test Sets}} & \multicolumn{1}{c|}{\textbf{Task 1}} & \multicolumn{2}{c|}{\textbf{Task 2}} & \multicolumn{1}{c|}{\textbf{Task 1}} & \multicolumn{2}{c|}{\textbf{Task 2}} & \multicolumn{1}{c|}{\textbf{Task 1}} & \multicolumn{2}{c|}{\textbf{Task 2 }} & \multicolumn{1}{c|}{\textbf{Task 1}} & \multicolumn{1}{c|}{\textbf{Task 2}} \\ \hline
\multicolumn{1}{|l|}{\multirow{2}{*}{\textbf{Datasets}}} & \multicolumn{1}{c|}{\textbf{\begin{tabular}[c]{@{}c@{}}Proprietary\end{tabular}}} & \multicolumn{2}{c|}{\textbf{\begin{tabular}[c]{@{}c@{}}Clarkson\end{tabular}}} & \multicolumn{1}{c|}{\textbf{\begin{tabular}[c]{@{}c@{}}Proprietary\end{tabular}}} & \multicolumn{2}{c|}{\textbf{\begin{tabular}[c]{@{}c@{}}Warsaw\end{tabular}}} & \multicolumn{1}{c|}{\textbf{\begin{tabular}[c]{@{}c@{}}Proprietary\end{tabular}}} & \multicolumn{2}{c|}{\textbf{\begin{tabular}[c]{@{}c@{}}Notre Dame\end{tabular}}} & \multicolumn{1}{c|}{\textbf{\begin{tabular}[c]{@{}c@{}}Proprietary\end{tabular}}} & \textbf{\begin{tabular}[c]{@{}c@{}}IIITD-WVU \end{tabular}} \\ \cline{2-12} 
\multicolumn{1}{|l|}{} & \multicolumn{1}{c|}{Test} & \multicolumn{2}{c|}{Test} & \multicolumn{1}{c|}{Test} & \multicolumn{1}{c|}{K. Test} & \multicolumn{1}{c|}{U. Test} & \multicolumn{1}{c|}{Test} & \multicolumn{1}{c|}{K. Test} & \multicolumn{1}{c|}{U. Test} & \multicolumn{1}{c|}{Test} & Test \\ \hline
\multicolumn{1}{|l|}{Task 1 Expert Model} & \multicolumn{1}{c|}{98.44} & \multicolumn{2}{c|}{28.63} & \multicolumn{1}{c|}{98.44} & \multicolumn{1}{c|}{92.95} & \multicolumn{1}{c|}{98.56} & \multicolumn{1}{c|}{98.44} & \multicolumn{1}{c|}{93.55} & \multicolumn{1}{c|}{91.00} & \multicolumn{1}{c|}{98.44} & 42.91 \\ \hline
\multicolumn{1}{|l|}{Task 2 Expert Model} & \multicolumn{1}{c|}{25.54} & \multicolumn{2}{c|}{92.05} & \multicolumn{1}{c|}{0.31} & \multicolumn{1}{c|}{100} & \multicolumn{1}{c|}{100} & \multicolumn{1}{c|}{29.90} & \multicolumn{1}{c|}{100} & \multicolumn{1}{c|}{66.55} & \multicolumn{1}{c|}{0.31} & 29.30 \\ \hline
\multicolumn{1}{|l|}{Fine-Tuned} & \multicolumn{1}{c|}{86.91} & \multicolumn{2}{c|}{93.51} & \multicolumn{1}{c|}{45.48} & \multicolumn{1}{c|}{100} & \multicolumn{1}{c|}{100} & \multicolumn{1}{c|}{98.75} & \multicolumn{1}{c|}{100} & \multicolumn{1}{c|}{99.77} & \multicolumn{1}{c|}{83.17} & 48.85 \\ \hline
\multicolumn{1}{|l|}{Full-Retrain} & \multicolumn{1}{c|}{96.57} & \multicolumn{2}{c|}{91.63} & \multicolumn{1}{c|}{93.76} & \multicolumn{1}{c|}{100} & \multicolumn{1}{c|}{100} & \multicolumn{1}{c|}{96.57} & \multicolumn{1}{c|}{100} & \multicolumn{1}{c|}{100} & \multicolumn{1}{c|}{96.57} & \textbf{66.81} \\ \hline
\multicolumn{12}{|c|}{\textbf{Ensemble of Task 1 and 2 Expert Models}} \\ \hline
\multicolumn{1}{|l|}{Equal Membership} & \multicolumn{1}{c|}{97.50} & \multicolumn{2}{c|}{89.67} & \multicolumn{1}{c|}{97.81} & \multicolumn{1}{c|}{99.45} & \multicolumn{1}{c|}{\textbf{100}} & \multicolumn{1}{c|}{\textbf{99.37}} & \multicolumn{1}{c|}{99.88} & \multicolumn{1}{c|}{96.22} & \multicolumn{1}{c|}{\textbf{98.44}} & 43.62 \\ \hline
\multicolumn{1}{|l|}{Pre-trained ViT-DM} & \multicolumn{1}{c|}{98.13} & \multicolumn{2}{c|}{72.80} & \multicolumn{1}{c|}{91.27} & \multicolumn{1}{c|}{\textbf{100}} & \multicolumn{1}{c|}{99.38} & \multicolumn{1}{c|}{99.37} & \multicolumn{1}{c|}{\textbf{100}} & \multicolumn{1}{c|}{80.44} & \multicolumn{1}{c|}{88.16} & 29.27 \\ \hline
\multicolumn{1}{|l|}{FE-DM (Proposed Method)} & \multicolumn{1}{c|}{\textbf{98.44}} & \multicolumn{2}{c|}{\textbf{92.67}} & \multicolumn{1}{c|}{\textbf{98.13}} & \multicolumn{1}{c|}{\textbf{100}} & \multicolumn{1}{c|}{\textbf{100}} & \multicolumn{1}{c|}{\textbf{99.37}} & \multicolumn{1}{c|}{\textbf{100}} & \multicolumn{1}{c|}{\textbf{99.55}} & \multicolumn{1}{c|}{98.13} & 44.94 \\ \hline
\end{tabular}
}
\vspace{-10pt}
\end{table*}

To evaluate the proposed method, we conduct experiments across three setups: LivDet-Iris-2017, LivDet-Iris-2020, and Split MNIST. The LivDet-Iris-2017 and LivDet-Iris-2020 setups involve two tasks in sequence, while the Split MNIST setup involves five tasks in sequence. The LivDet-Iris-2017 setup represents a scenario where a distribution shift occurs between tasks, but no shift occurs within a task (except in one case, explained in Section \ref{sec:LivDet-Iris-2017Setup}). The LivDet-Iris-2020 setup illustrates a scenario where distribution shifts occur both between and within tasks. The Split MNIST setup represents a scenario where the distributions of different tasks are disjoint, but there is no shift within tasks. The LivDet-Iris-2017 and LivDet-Iris-2020 setups reflect practical scenarios in the application of presentation attack (PA) detection, i.e., spoof detection, for iris biometric recognition systems. In this case, PA detection is treated as a binary classification problem, distinguishing between bonafide iris images and PA images (such as prints, cosmetic contact lenses, artificial eyes, and electronic displays). The Split MNIST setup is a simulated continuous learning scenario, used to compare the proposed method with existing state-of-the-art (SOTA) continuous learning techniques.

For training the feature extractor (FE) model, we initialize the weights from a pre-trained ViT-Base model \cite{Dosovitskiy2021} trained on the ImageNet-21k and JFT-300M datasets. We remove the MLP head used for classification from the original ViT architecture to make it a feature extractor. 
We use 100 epochs, a batch size of 15, 0.25 as the value of $\tau$ and the stochastic gradient descent (SGD) optimizer with a learning rate of 1e-5. For the implementation of the LOF distance measure, we use the default values provided in \cite{Breunig2000}, 20 as the number of neighbors, and the Euclidean distance as the distance metric.

\subsection{LivDet-Iris-2017 Setup and Results}
\label{sec:LivDet-Iris-2017Setup}
In this setup, we use two datasets: a proprietary dataset and the LivDet-Iris-2017 dataset \cite{Yambay2017}. The proprietary dataset is used for Task 1, and the LivDet-Iris-2017 dataset is used for Task 2. The proprietary dataset and expert models are taken from \cite{Sharma2020}. The LivDet-Iris-2017 dataset \cite{Yambay2017} is a publicly available dataset for iris presentation attack (PA) detection. It consists of four subsets: Clarkson, Warsaw, Notre Dame, and IIITD-WVU. Each subset contains corresponding training and test sets. The Warsaw and Notre Dame test sets are further split into known and unknown test splits based on types of PAs included in the test set with respect to the training set. The Clarkson and Notre Dame test sets correspond to the cross-PA scenario, while the Warsaw data represent a cross-sensor scenario. The IIITD-WVU subset represents a cross-dataset scenario where a distribution shift also occurs within the task. 
Table \ref{table:LivDet-Iris-2017-Dataset} summarizes the training and test sets for both tasks, along with the number of images present in each set.
\setcounter{figure}{2}
\begin{figure*}
     \centering
     \begin{subfigure}[b]{0.8\columnwidth}
         \centering
         \includegraphics[scale=0.3]{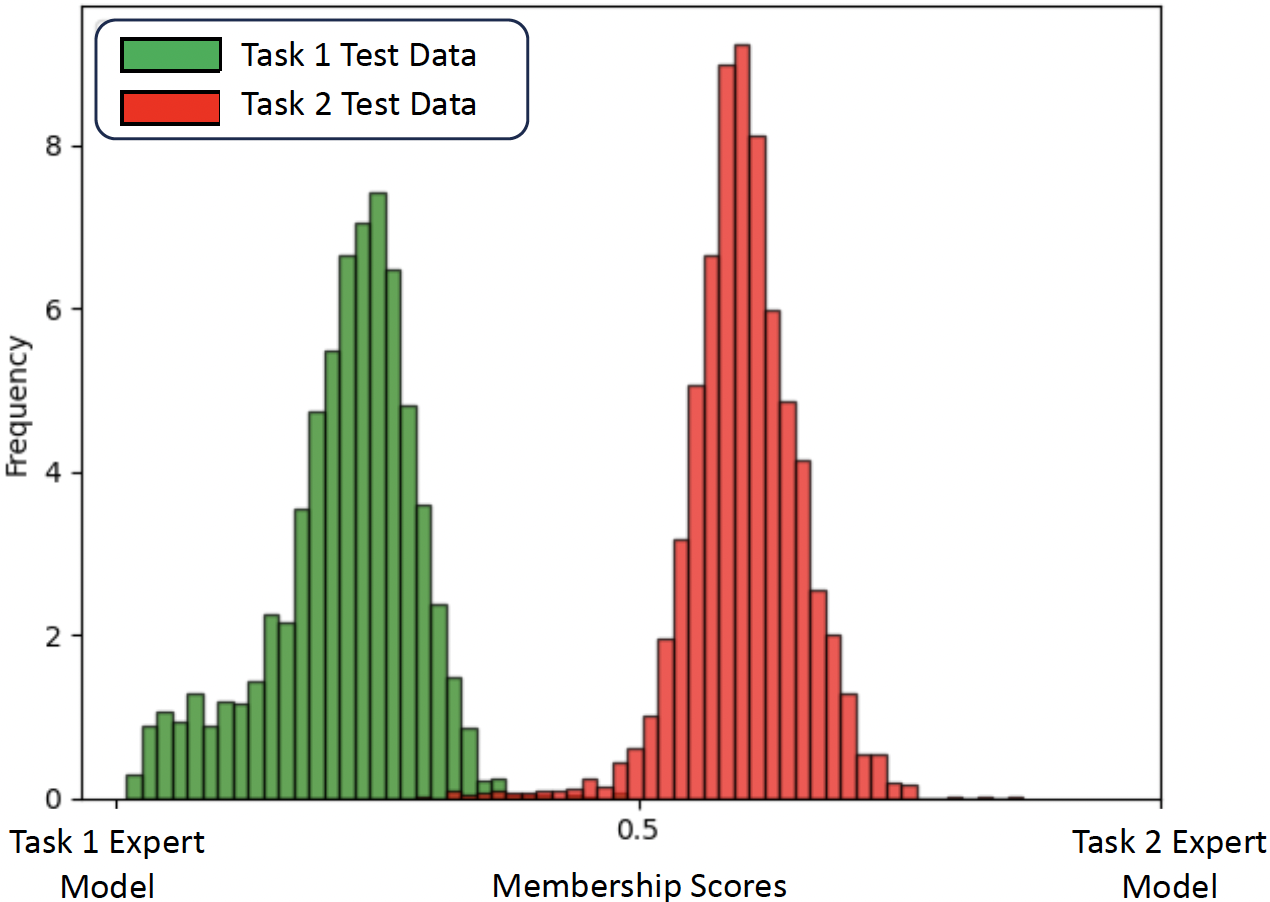}
         \caption{}
     \end{subfigure}
     \begin{subfigure}[b]{0.8\columnwidth}
         \centering
         \includegraphics[scale=0.3]{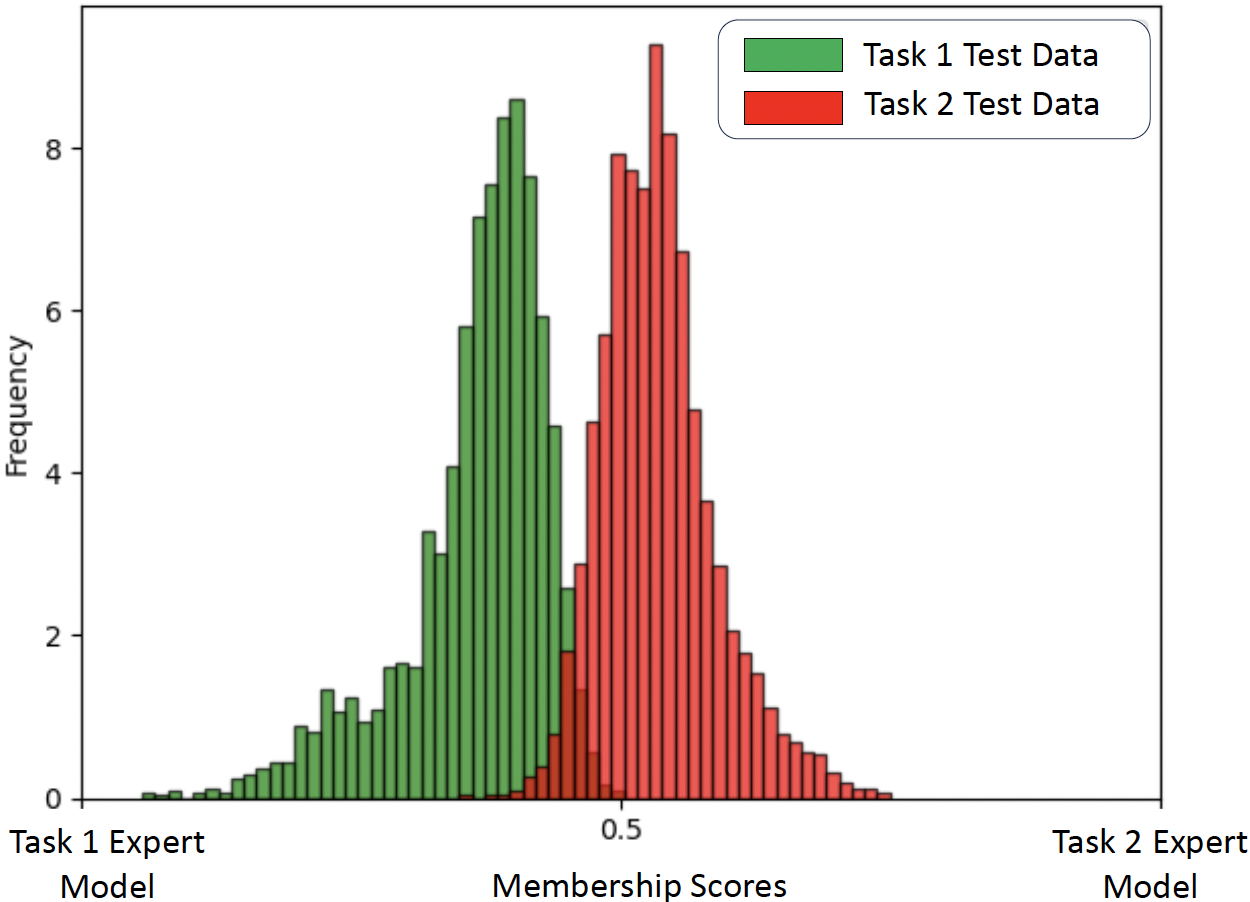}
         \caption{}
     \end{subfigure}
     \begin{subfigure}[b]{0.8\columnwidth}
         \centering
         \includegraphics[scale=0.3]{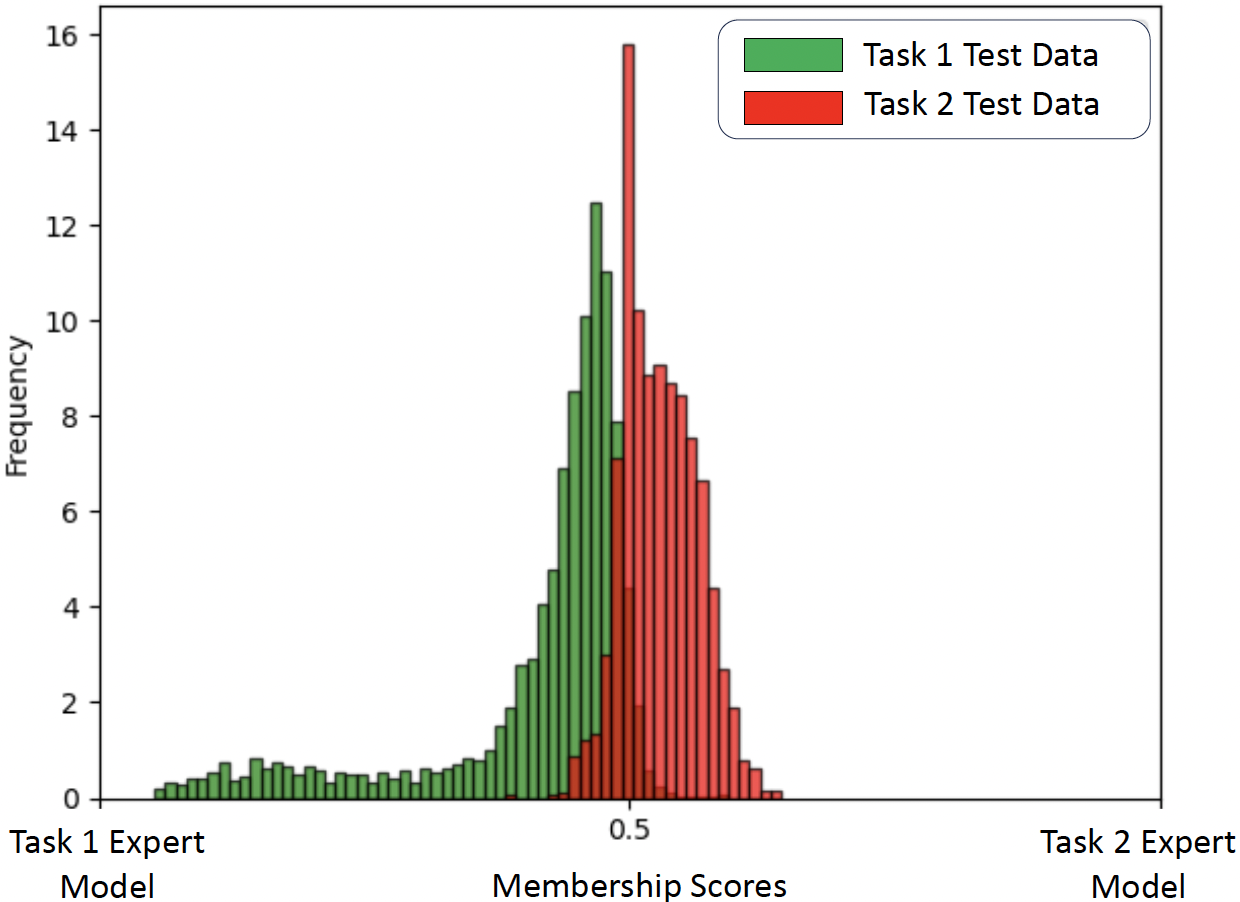}
         \caption{}
     \end{subfigure}
     \begin{subfigure}[b]{0.8\columnwidth}
         \centering
         \includegraphics[scale=0.3]{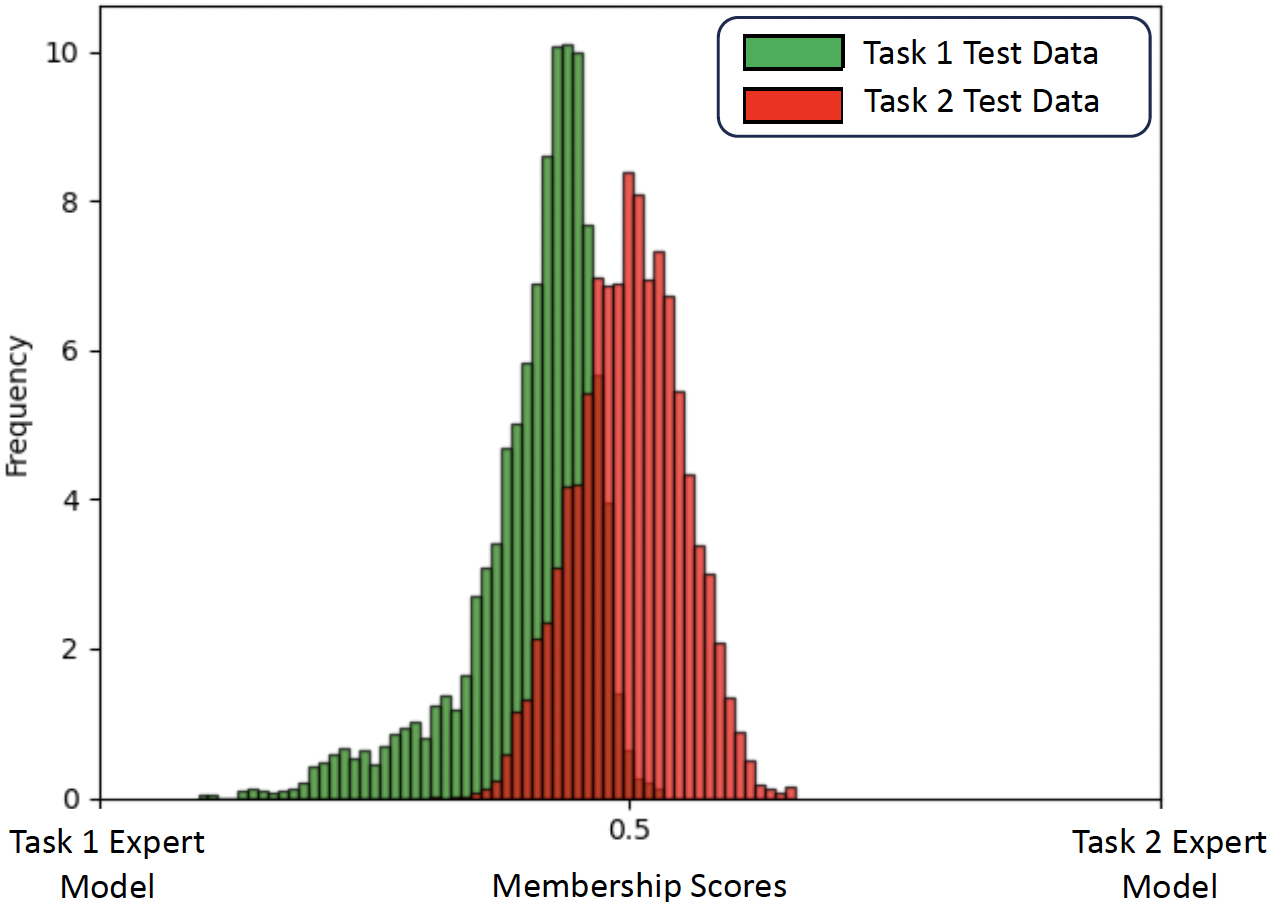}
         \caption{}
     \end{subfigure}
     \caption{The histogram of membership scores assigned to all test samples (tasks 1 and 2) corresponding to (a) Clarkson, (b) Warsaw, (c) Notre Dame, and (d) IIIT-WVU subsets of the LivDet-Iris-2017 setup. In the case of Warsaw and Notre Dame, `Known' test splits are used for illustration. Membership values toward `0' on the x-axis symbolize higher priority given to the task 1 expert model, whereas membership values toward `1' on the x-axis denote higher priority given to the task 2 expert model. The figure is better viewed in color.} 
     \label{fig:LivDet-Iris-2017-Weights}
     \vspace{-10pt}
\end{figure*}

For comparison, we use the following methods: (i) \textbf{Task 1 Expert Model}: trained only on Task 1, (ii) \textbf{Task 2 Expert Model}: trained only on Task 2, (iii) \textbf{Fine-Tuned}: trained on Task 1 and then fine-tuned on Task 2, (iv) \textbf{Full-Retrain}: trained on both Task 1 and Task 2, (v) \textbf{Equal Membership}: ensemble of both expert models with equal membership, (vi) \textbf{Pre-trained ViT-DM}: ensemble of both expert models with dynamic membership, where the in-domain model uses a pre-trained ViT model for feature representation, and (vii) \textbf{FE-DM (Proposed Method)}: ensemble of both expert models with dynamic membership, where the proposed feature extractor (FE) model is used to represent the task data. Table \ref{table:LivDet-Iris-2017-Results} presents the performance of all the models in terms of True Detection Rate (TDR(\%)) at a 0.2\% False Detection Rate (FDR), as commonly used in the iris PA detection literature \cite{Sharma2020}. TDR represents the percentage of PA samples correctly detected, while FDR denotes the percentage of bonafide samples incorrectly detected as PA. Performance scores are reported individually for both Task 1 and Task 2 test splits. The goal is to achieve high performance (higher TDR) on both splits. The Full-Retrain model serves as an upper benchmark for evaluating the performance of continuous learning methods.

The Task 1 and Task 2 expert models perform well on their respective test splits but fail on the other task\textquotesingle s test split. The Fine-Tuned model performs well on Task 2 but suffers from catastrophic forgetting, resulting in poor performance on Task 1 test split. The Full-Retrain model performs well on both test splits but is not a practical solution, as old training data is often unavailable in real-world scenarios. For the ensemble methods, we use the Equal Membership method to highlight the importance of dynamic membership and the Pre-trained ViT-DM method to demonstrate the relevance of the proposed FE module. The proposed method outperforms Full-Retrain (except for the IIIT-WVU test split) and both ensemble-based approaches, validating that the proposed FE model better represents the task data and that memberships are appropriately assigned to their respective expert models. We also visualize membership histograms for various test splits (Figure \ref{fig:LivDet-Iris-2017-Weights}). These histograms show the membership scores assigned to the Task 2 expert model from both tasks\textquotesingle~ test data. So membership scores closer to 0 indicate a higher priority for the Task 1 expert model, while scores closer to 1 indicate a higher priority for the Task 2 expert model. In all cases, the Task 1 test set receives higher scores for the Task 1 expert model, and the Task 2 test set receives higher scores for the Task 2 expert model, except for the IIIT-WVU split. The membership scores for the IIIT-WVU test set are around 0.5, as the distribution of the IIIT-WVU test set is independent of the training distributions of both tasks, which is expected behavior. 

\subsection{LivDet-Iris-2020 Setup and Results}
In this setup, we utilize three datasets: the proprietary dataset, the Warsaw PostMortem v3 dataset \cite{WarsawPostmortem}, and the LivDet-Iris-2020 dataset \cite{Das2020}. We divide the proprietary dataset into three splits: one for the Task 1 training set and two for the Task 2 training sets. The Warsaw PostMortem v3 dataset is used as the third training split for Task 2, while the LivDet-Iris-2020 dataset serves as the test set, which exhibits a distribution shift from both Task 1 and Task 2 training sets. The different training splits for Task 2 correspond to no distribution shift, a cross-sensor shift, and a cross-PA shift. Table \ref{table:LivDet-Iris-2020-Dataset} details the number of bonafide and PA images used in these sets. Table \ref{table:LivDet-Iris-2020-Results} presents the performance of all the models in terms of TDR at a 0.2\% False Detection Rate (FDR) on the LivDet-Iris-2020 test dataset.

\begin{table*}[]
\captionsetup{font=small}
\caption{Description of the task 1 and 2 train/test sets in the LivDet-Iris-2020 setup, along with the number of bonafide and PA iris images present in the sets.}
\label{table:LivDet-Iris-2020-Dataset}
\resizebox{\textwidth}{!}{%
\begin{tabular}{|l|c|cccc|c|}
\hline
\textbf{\large Domains} & \textbf{\large Task 1 Train Set} & \multicolumn{4}{c|}{\textbf{\large Task 2 Train Set}} & \textbf{\large Test Set (Task 1 and 2)} \\ \hline
\textbf{\large Datasets} & \textbf{\large Proprietary Split I} & \multicolumn{1}{c|}{\textbf{\large Proprietary Split II}} & \multicolumn{1}{c|}{\textbf{\large Proprietary Split III}} & \multicolumn{1}{c|}{\textbf{\large Warsaw PostMortem v3}} & \textbf{\large Combined} & \textbf{\large LivDet-Iris 2020} \\ \hline
\textbf{\large Train/Test} & \textbf{\large Train} & \multicolumn{1}{c|}{\textbf{\large Train}} & \multicolumn{1}{c|}{\textbf{\large Train}} & \multicolumn{1}{c|}{\textbf{\large Train}} & \textbf{\large Train} & \textbf{\large Test} \\ \hline
\large Bonafide & \large 9,660 & \multicolumn{1}{c|}{\large 2,963} & \multicolumn{1}{c|}{\large 9,606} & \multicolumn{1}{c|}{-} & \large 12,569 & \large 5,331 \\ \hline
\large PA & \large 6,075 & \multicolumn{1}{c|}{\large 352} & \multicolumn{1}{c|}{\large 922} & \multicolumn{1}{c|}{\large 2,400} & {\large 3,674} & \large 7,101 \\ \hline
\end{tabular}
}
\vspace{-5pt}
\end{table*}

\begin{table*}[]
\captionsetup{font=small}
\caption{The performance of all methods in terms of True Detection Rate (\%, higher the better) at 0.2\% False Detection Rate on the LivDet-Iris-2020 test set.}
\label{table:LivDet-Iris-2020-Results}
\centering
\resizebox{\textwidth}{!}{%
\begin{tabular}{|lccccc|}
\hline
\multicolumn{1}{|l|}{\textbf{Domains}} & \multicolumn{1}{l|}{\textbf{Task 1}} & \multicolumn{4}{c|}{\textbf{Task 2}} \\ \hline
\multicolumn{1}{|l|}{\textbf{Train Dataset}} & \multicolumn{1}{l|}{\textbf{Proprietary  Split I}} & \multicolumn{1}{l|}{\textbf{Proprietary  Split II}} & \multicolumn{1}{l|}{\textbf{Proprietary  Split III}} & \multicolumn{1}{l|}{\textbf{Warsaw PostMortem v3}} & \multicolumn{1}{l|}{\textbf{Combined}} \\ \hline
\multicolumn{1}{|l|}{\textbf{Test Dataset}} & \multicolumn{5}{c|}{\textbf{LivDet-Iris 2020 (Task 1 and 2 Test Set)}} \\ \hline
\multicolumn{1}{|l|}{Task 1 and 2 Expert Models} & \multicolumn{1}{c|}{61.86} & \multicolumn{1}{c|}{58.25} & \multicolumn{1}{c|}{75.55} & \multicolumn{1}{c|}{0.94} & 85.56 \\ \hline
\multicolumn{1}{|l|}{Fine-Tuned} & \multicolumn{1}{c|}{-} & \multicolumn{1}{c|}{63.18} & \multicolumn{1}{c|}{66.53} & \multicolumn{1}{c|}{0} & 83.00 \\ \hline
\multicolumn{1}{|l|}{Full-Retrain} & \multicolumn{1}{c|}{-} & \multicolumn{1}{c|}{\textbf{77.96}} & \multicolumn{1}{c|}{76.96} & \multicolumn{1}{c|}{67.76} & \textbf{94.05} \\ \hline
\multicolumn{6}{|c|}{\textbf{Ensemble of Task 1 and 2 Expert Models}} \\ \hline
\multicolumn{1}{|l|}{Equal Membership} & \multicolumn{1}{c|}{-} & \multicolumn{1}{c|}{\textbf{72.42}} & \multicolumn{1}{c|}{79.04} & \multicolumn{1}{c|}{58.73} & 87.05 \\ \hline
\multicolumn{1}{|l|}{Pre-trained ViT-DM} & \multicolumn{1}{c|}{-} & \multicolumn{1}{c|}{69.91} & \multicolumn{1}{c|}{79.03} & \multicolumn{1}{c|}{58.73} & 89.38 \\ \hline
\multicolumn{1}{|l|}{FE-DM (Proposed Method)} & \multicolumn{1}{c|}{-} & \multicolumn{1}{c|}{69.27} & \multicolumn{1}{c|}{\textbf{81.36}} & \multicolumn{1}{c|}{\textbf{61.99}} & 93.62 \\ \hline
\end{tabular}
}
\vspace{-10pt}
\end{table*}

The Task 1 and Task 2 expert models perform poorly on the test set, as its distribution differs from the training sets of both tasks. A similar issue occurs with the Fine-Tuned model. While the Full-Retrain model outperforms the other models, it is impractical due to the unavailability of the old training data. The proposed method, however, outperforms both ensemble methods (Equal Membership and Pre-trained ViT-DM) and achieves comparable performance to the Full-Retrain model.

\subsection{Split MNIST Setup and Results}
\label{subsec:MNIST}
We also conduct experiments on the MNIST dataset to compare the proposed method with existing SOTA continuous learning strategies. The original dataset consists of 28 $\times$ 28 images of ten digits. We use the standard train-test split, with 60,000 training images and 10,000 test images. The primary task is to distinguish even-digit images from odd-digit images, which is subdivided into five binary sub-tasks. The first task classifies the digits `0' and `1', the second task classifies `2' and `3', and so on. This dataset split is known as Split MNIST in the literature \cite{Hsu2018,Vandeven2018}. The class labels remain consistent across all tasks. In this setup, the distributions of training data for different tasks are disjoint, but there is no shift between the training and test distributions within each task.

For a fair comparison, we use a multi-layer perceptron (MLP) architecture defined in \cite{Hsu2018} as the expert model. We compare the proposed method against Fine-Tuned, Full-Retrain, Equal Membership, Manual Membership, Pre-trained ViT-DM, and other SOTA continuous learning methods. In the Manual Membership approach, we manually assign a membership score of 1 to the correct expert model and 0 to the others. This method serves as an upper bound, as the training sets for all sub-tasks are disjoint. Table \ref{table:MNIST-Results} presents the results of all methods. 
\begin{table}[]
\captionsetup{font=small}
\caption{The average accuracy (\%, higher the better) of the proposed approach with different SOTA continuous learning methods on the Split MNIST dataset. Methods with `+' superscript are reported from \cite{Hsu2018}, `o' from \cite{Kao2021}, `*' from \cite{Bang2021} and `-' from \cite{Lee2020}. ARI \cite{Wang2022} performance is reported from their own paper. All methods utilize the same experimental setup and expert models but differ in hyperparameters (batch size, learning rate, and number of epochs). We use the same hyperparameters as used in \cite{Hsu2018}. Each value is an average of ten runs.}
\label{table:MNIST-Results}
\centering
\resizebox{0.9\columnwidth}{!}{%
\begin{tabular}{|ll|}
\hline
\multicolumn{1}{|l|}{\textbf{Method}} & \textbf{Accuracy (\%)} \\ \hline
\multicolumn{1}{|l|}{Fine-Tuned} & 63.20 ± 0.35 \\ \hline
\multicolumn{1}{|l|}{Full-Retrain} & 98.59 ± 0.15 \\ \hline
\multicolumn{1}{|l|}{EWC$^{+}$\cite{Kirkpatrick2017}} & 58.85 ± 2.59 \\ \hline
\multicolumn{1}{|l|}{Online EWC$^{+}$ \cite{Schwarz2018}} & 57.33 ± 1.44 \\ \hline
\multicolumn{1}{|l|}{SI$^{+}$ \cite{Zenke2017}} & 64.76 ± 3.09 \\ \hline
\multicolumn{1}{|l|}{KFAC$^{o}$ \cite{Ritter2018}} & 67.86 ± 1.33 \\ \hline
\multicolumn{1}{|l|}{MAS$^{+}$ \cite{Aljundi2018}} & 68.57 ± 6.85 \\ \hline
\multicolumn{1}{|l|}{LwF$^{+}$ \cite{Li2017}} & 71.02 ± 1.26 \\ \hline
\multicolumn{1}{|l|}{OWM$^{o}$ \cite{Zeng2019}} & 87.46 ± 0.74 \\ \hline
\multicolumn{1}{|l|}{NCL$^{o}$ \cite{Kao2021}} & 91.48 ± 0.64 \\ \hline
\multicolumn{1}{|l|}{BiC$^{*}$ \cite{Wu2019}} & 77.75 ± 1.27 \\ \hline
\multicolumn{1}{|l|}{ER$^{-}$ \cite{Chaudhry2019}} & 85.69 \\ \hline
\multicolumn{1}{|l|}{GDumb$^{*}$ \cite{Prabhu2020}} & 88.51 ± 0.52 \\ \hline
\multicolumn{1}{|l|}{RM$^{*}$ \cite{Bang2021}} & 92.65 ± 0.33 \\ \hline
\multicolumn{1}{|l|}{DGR$^{+}$ \cite{Shin2017}} & 95.74 ± 0.23 \\ \hline
\multicolumn{1}{|l|}{GEM$^{+}$ \cite{Lopez2017}} & 96.16 ± 0.35 \\ \hline
\multicolumn{1}{|l|}{RtF$^{+}$ \cite{Van2018}} & 97.31 ± 0.11 \\ \hline
\multicolumn{1}{|l|}{ARI \cite{Wang2022}} & \textbf{98.91} \\ \hline
\multicolumn{2}{|l|}{\textbf{Ensemble of Expert Models}} \\ \hline
\multicolumn{1}{|l|}{Equal Membership (Lower Limit)} & 84.20 ± 0.08 \\ \hline
\multicolumn{1}{|l|}{Manual Membership (Upper Limit)} & 98.66 ± 0.008 \\ \hline
\multicolumn{1}{|l|}{CN-DPM$^{-}$ \cite{Lee2020}} & 93.23 \\ \hline
\multicolumn{1}{|l|}{Pre-trained ViT-DM} & 81.34 ± 0.005 \\ \hline
\multicolumn{1}{|l|}{FE-DM (Proposed Method)} & 94.32 ± 0.01 \\ \hline
\multicolumn{1}{|l|}{\begin{tabular}[c]{@{}l@{}}FE-DM with Mahalanobis Distance\\ (Proposed Method)\end{tabular}} & \textit{97.03 ± 0.0001} \\ \hline
\end{tabular}
}
\vspace{-10pt}
\end{table}
The proposed method outperforms Fine-Tuned, ensemble-based methods, and other SOTA approaches. However, its performance is slightly lower than that of four replay-based methods: DGR \cite{Shin2017}, RtF \cite{Van2018}, GEM \cite{Lopez2017}, and ARI \cite{Wang2022}. DGR \cite{Shin2017}, RtF \cite{Van2018}, and ARI \cite{Wang2022} are generative-based methods that involve training a separate generative model which is further used to augment the training of subsequent tasks. While this process improves performance, it also increases training time and makes the expert model dependent on the generative model. GEM \cite{Lopez2017} and ARI \cite{Wang2022} methods also require additional memory to store a subset of the previous task samples, which raises concerns about both memory usage and privacy. ARI \cite{Wang2022} further utilizes task identity during training. \textbf{However, the proposed method does not generate previous task samples, does not rely on task identity, and keeps the expert models independent from additional models.} The Manual Membership method achieves the highest performance ($98.66\%$), surpassing all other methods, including the Full-Retrain method, and is comparable to ARI \cite{Wang2022} ($98.91\%$), highlighting the potential of ensemble-based approaches. When we experiment with an alternative distance measure, viz., Mahalanobis distance, it yields an accuracy of $97.03\%$, which is comparable to the highest performance. In this setup, Mahalanobis distance performs best, as the disjoint training distributions are effectively captured by the mean and variance, whereas LOF outperforms in the other setups. We also evaluate the forgetting behavior of our proposed approach (LOF as distance measure) using the Backward Transfer (BWT) metric \cite{Lopez-Paz2017}, considering the sequential learning of tasks on Split MNIST dataset. We achieve a BWT of +0.9\%. Typically, BWT is negative, indicating the extent of forgetting in previous tasks. However, the positive value in our case, suggests minimal forgetting, which could be attributed to our multi-model design that maintains independence across expert models.      

\begin{figure}[t]
    \centering
    \begin{subfigure}[t]{0.5\columnwidth}
        \centering
        \includegraphics[scale=0.35]{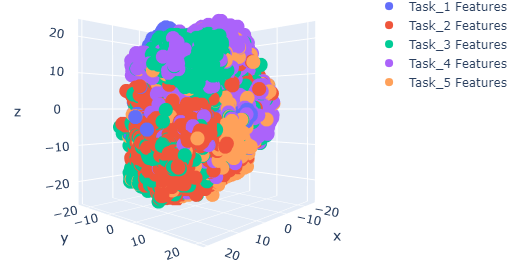}
        \caption{}
        \label{fig:ViT_PreTrained_3D}
    \end{subfigure}%
    \begin{subfigure}[t]{0.5\columnwidth}
        \centering
        \includegraphics[scale=0.35]{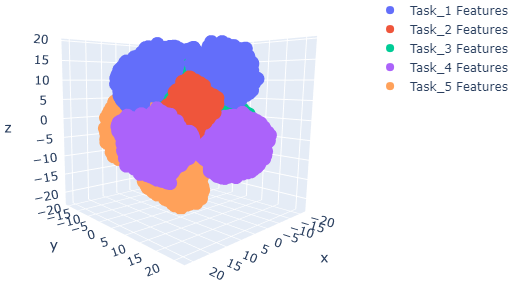}
        \caption{}
        \label{fig:ViT_FineTuned_3D}
    \end{subfigure}
    \caption{3-D t-sne plots correspond to five sub-tasks of the Split MNIST dataset using (a) pre-trained ViT and (b) our trained ViT embeddings. Pre-trained ViT embeddings of different classes overlap with each other, whereas trained ViT embeddings form clusters of different classes. The figure is better viewed in color.}
    \vspace{-10pt}
\end{figure}

To further emphasize the importance of our proposed FE module, we visualize the features extracted from pre-trained ViT model (Figure \ref{fig:ViT_PreTrained_3D}) and our trained FE model (Figure \ref{fig:ViT_FineTuned_3D}) using t-SNE \cite{Van2008} (three dimensions). The features extracted by the pre-trained model exhibit significant overlap among the embeddings of different tasks, in contrast to the embeddings from our trained FE model, which are more distinct. Both the experimental results and the visualizations validate the effectiveness of the proposed loss functions in training the FE model.

\subsection{Findings}
The main findings from the three experimental setups are as follows:

\noindent 1. When test distribution is similar to training data, the proposed method outperforms all the methods, including Full-Retrain, as shown in the LivDet-Iris-2017 setup (Table \ref{table:LivDet-Iris-2017-Results}).

\noindent 2. When there is a shift in test data distribution relative to all tasks\textquotesingle~ training data, the proposed approach still outperforms other methods and is comparable to the Full-Retrain method, as shown in the LivDet-Iris-2020 setup (Table \ref{table:LivDet-Iris-2020-Results}). 

\noindent 3. In cases where the training distributions of different tasks are disjoint, the proposed approach outperforms Fine-Tuned, ensemble-based approaches, and various other SOTA methods. Its performance is lower than some of the replay-based methods, which could be improved using the Mahalanobis distance as a distance measure.

\noindent 4. The proposed in-domain model effectively assigns membership scores to the respective expert models, leading to superior performance compared to the Equal Membership method as evident from the result in Tables \ref{table:LivDet-Iris-2017-Results}, \ref{table:LivDet-Iris-2020-Results}, and \ref{table:MNIST-Results}. The membership histograms in Figure \ref{fig:LivDet-Iris-2017-Weights} further validate the accurate allocation of membership scores.

\noindent 5. The proposed FE model better represents the training data, as shown by its superior performance compared to the Pre-trained ViT-DM model in Tables \ref{table:LivDet-Iris-2017-Results}, \ref{table:LivDet-Iris-2020-Results}, and \ref{table:MNIST-Results}. The visual representation in Figure \ref{fig:ViT_FineTuned_3D} further supports this finding.

\section{Discussion on Memory Requirement and Scalability}
\label{sec:Discussion}

Regarding memory requirements, the proposed method is more memory-efficient compared to other approaches, especially replay-based methods, as it does not have to store or transfer previous task images or features to subsequent tasks. Instead, the proposed method stores previous tasks\textquotesingle~ information in in-domain models (requires much lower memory than images/features). However, this setup raises concerns about scalability, when the number of models increases linearly with the number of tasks. To address this, the number of models can be reduced in several ways: (i) apply preconditions (such as performance or distribution shift measures) before building additional models; (ii) select a subset of expert/in-domain models based on prior knowledge of the test data; or (iii) merge expert and in-domain models using techniques like knowledge distillation \cite{Braun2024, Liu2021} or other methods \cite{Singh2020, Stoica2024}.

To address the scalability issue, we conducted a small experiment where we merged expert models using knowledge distillation \cite{Hinton2014}. For this experiment, we considered the first three consecutive sub-tasks of the Split MNIST dataset (Section \ref{subsec:MNIST}). We trained a student expert model based on two teacher expert models (tasks 1 and 2), each separately trained on the first two tasks in the sequence. The student expert model uses the same architecture as the teacher models and is trained without access to the original training data. To achieve this, we generate synthetic data approximating the data distributions on which the teacher models were initially trained, using Gaussian random variables.
Next, we combine two in-domain models (tasks 1 and 2) into a single in-domain model, averaging their outputs with a weighting factor of 0.5. These fused models, which capture the knowledge from tasks 1 and 2, are then evaluated alongside models trained on task 3. The results show that our proposed method achieves an accuracy of 90.9\% across these three tasks. Notably, when we combine the models of tasks 1 and 2, the accuracy increases to 93.1\%. This experiment demonstrates that the approach is potentially scalable and effective for handling multiple tasks. It not only reduces the number of models that need to be maintained, but also improves overall performance.

\section{Summary}
\label{sec:Conclusion}

We propose a task-conditioned ensemble-based method for continuously learning an existing expert model. The method introduces an in-domain model that provides membership information to dynamically combine different task expert models. Evaluation of the proposed approach across three experimental setups, each representing different levels of distribution shifts, demonstrates its effectiveness. Since the method does not alter the existing expert models—either through the training process or by adding new architecture—it facilitates the reuse of these models. In future work, we plan to apply this approach to other application areas.

\balance

{
    \small
    \bibliographystyle{ieeenat_fullname}
    \bibliography{main}
}


\end{document}